# Exponential discretization of weights of neural network connections in pre-trained neural networks

*Malsagov M.Yu.[1], Khayrov E.M.[1], Pushkareva M.M.[1], Karandashev I.M.[1,2]*

[1] Center of Optical Neural Technologies, Scientific Research Institute for System Analysis RAS, Moscow, Nakhimovskiy prosp., 36, b.1., 117218, Russia
[2] Peoples Friendship University of Russia (RUDN University), 6 Miklukho-Maklaya St, Moscow, 117198, Russian Federation
malsagov@niisi.ras.ru, emil.khayrov@gmail.com, mariaratko@gmail.com, karandashev@niisi.ras.ru

**Abstract.** To reduce random access memory (RAM) requirements and to increase speed of recognition algorithms we consider a weight discretization problem for trained neural networks. We show that an exponential discretization is preferable to a linear discretization since it allows one to achieve the same accuracy when the number of bits is 1 or 2 less. The quality of the neural network VGG-16 is already satisfactory (top5 accuracy 69%) in the case of 3 bit exponential discretization. The ResNet50 neural network shows top5 accuracy 84% at 4 bits. Other neural networks perform fairly well at 5 bits (top5 accuracies of Xception, Inception-v3, and MobileNet-v2 top5 were 87%, 90%, and 77%, respectively). At less number of bits, the accuracy decreases rapidly.

1. **Introduction**

Present-day neural networks include millions parameters and their sizes become larger and larger. This is a global problem preventing deployment of neural networks on mobile devices and different chips. For example, even a relatively small pre-trained neural network VGG16 [1] designed to image classification [2] occupies about 553 MB in the device's memory.

There are different methods allowing us to decrease the size of the pre-trained neural network. In particular, one of these methods is a discretization that is a reduction of bit width of the neural network weights by dividing the weight distribution interval into discrete values. Applying the discretization to pre-trained neural networks, we significantly decrease the sizes of the weights and "lighten" the neural network making it applicable for realization and launch on mobile devices. In the papers [3-6], they examined clipping operations and quantization of matrix weights of neural networks and obtained analytic estimates for an optimal discrete representation for the Hopfield neural network weights as well as formulas for probabilities of errors in the obtained model. For the Hopfield model based on binary neurons, it was shown that the optimal discretization of the connection weights corresponded to maximization of correlations between the input and the discretized weights. In the present paper, we use the same idea, but for feedforward deep neural networks.

The most realizations of discretization of neural networks weights include the neural networks re-training in the course of the discretization; however, this process requires a huge computing power [7-9]. In contrast to the cited works, we examine discretization methods where the re-training is not necessary. In other words, we analyze the discretization of the weights of already trained neural networks. In the framework of such approach, the huge computing power is not necessary. The similar method was discussed in paper [10] where the authors proposed an approach known as a logarithmic discretization. They partitioned the weights into groups where the values of the logarithm to base 2 of the weights were equal to an integer from the interval [-7; 0]. In the present paper, we propose a more general approach in the framework of which we can vary the base of logarithm as well as the initial value. We called it an exponential discretization.

## 2. Description of discretization procedure

Let us assume initially that the neural network weights we have to discretize are distributed symmetrically over the interval $[-M, M]$, where $M$ is maximum absolute value of the weights. When we keep the signs of the weights in a separate array (providing one bit for the sign) absolute values of the numbers are distributed over the range $[0, M]$. For simplicity, we divide all of them by $M$. Consequently, in what follows that all the numbers are distributed over the interval $x \in [0,1]$.

When discretizing our goal is to decrease the variety of numbers by means of dividing the interval $x \in [0,1]$ into $n$ intervals whose ends are at the points

$$0 < x_0 < x_1 < x_2 < ... < x_{n-2} < x_{n-1} = 1 \qquad (1)$$

Note, the first interval is $[0, x_0]$ and the last interval is $[x_{n-2}, 1]$; all the other intervals are defined as $[x_{k-1}, x_k]$, $k = 1,...,n-1$. The number of bits $B$ necessary to store the discretized weight define the number of intervals $n$:

$$n = 2^{B-1}; \qquad (2)$$

at that we hold one bit for the sign of the number.

Although an arbitrary partition into intervals (1) is possible, we examined the two most popular ways only: the linear and the exponential partition (see Fig. 1).

### 2.1. Linear partition
In this case, we divide the interval of distribution into the same intervals:

$$x_k = x_0 + kq, \quad x_k = x_{k-1} + q, \quad k = 0,...,n-1 \qquad (3)$$

Here $x_0$ is the first point (a parameter of variation), $n$ is the number of intervals, and the length of the intervals $q$ we determine from the equality $x_{n-1} = 1$:

$$q = \frac{1 - x_0}{n - 1}. \qquad (4)$$

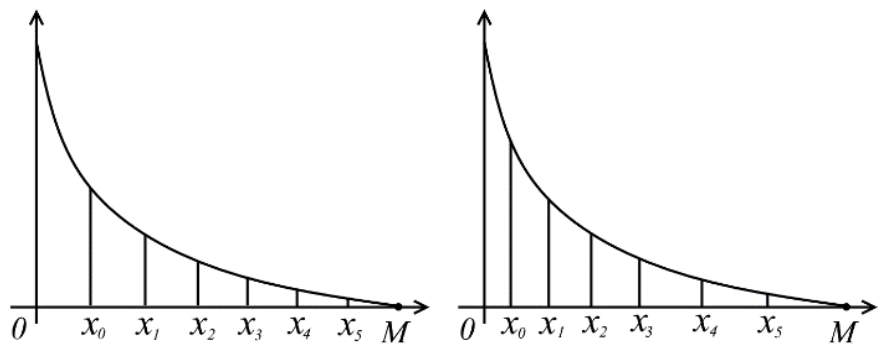

**Fig.1.** Equidistant linear distribution (left panel) and exponential distribution (right panel).

*2.2. Exponential partition*

In the case of the exponential partition, the subsequent intervals increase exponentially with respect of first interval:

$$x_k = x_0 q^k, \quad x_k = q x_{k-1}, \quad k = 0,...,n-1 \qquad (5)$$

As before, the first point $x_0$ is the parameter of variation and from the equality $x_{n-1} = 1$ we derive the formula for the length of the intervals

$$q = 1/\sqrt[n-1]{x_0}. \qquad (6)$$

*2.3. The value of $x_0$*

When implementing the both linear and exponential discretization algorithms it is necessary to define the initial number $x_0$. We do this maximizing the correlation between the 32-bit (the full precision) weights and the discretized weights. Let $x$ be the initial values of the weights in a layer and let $y$ be the discretized values. The expression for the correlation has the form

$$\rho(x, y) = \frac{\overline{xy} - \overline{x}\,\overline{y}}{\sigma_x \sigma_y} \qquad (7)$$

Optimal value $x_0$ corresponds to the maximal correlation. We suppose that the higher correlation between the discretized and initial weights the less the error in the work of the trained network.

*2.4. Rounding of numbers*

For a given number of bits $B$ the discretization procedure is defined by the position of the end of the initial interval (the value of $x_0$) as well as by the rounding method. There are different variants of rounding of numbers. We examined three of them: the floor, the ceiling, and the mean method.

When we use the rounding *floor* we replace all the numbers inside the interval $[x_i, x_{i+1}]$ by the lower boundary of the interval $x_i$.

When we use the rounding *ceil* we replace all the numbers inside the interval by the upper boundary of the interval $x_{i+1}$.

In the case of the rounding *mean,* we replace all the numbers inside the interval by the mean value of the numbers inside this interval.

In Fig. 2 we show the full code of the discretization procedure.

```
def quantize(W, x_0, B):
    M = np.fabs(W).max()
    n = 2^{B-1}
    signs = np.sign(W)
    x = np.fabs(W.flatten()) / M      #   normalized values between 0 and 1
    q = x_0^{-1/(n-1)}  # for exponential
#   q = (1 - x_0) / (n - 1) # for linear
    xx = [0] + [x0 * q ** i for i in range(n)]  # for exponential
#   xx = [0] + [x0 + q * i for i in range(n)]   # for linear
    W_q = np.zeros(W.size, dtype=np.float)
    for i in range(len(xx) - 2):
        ii = np.logical_and(xx[i] < x, x <= xx[i +1])
        W_q[ii] = np.mean(W_{norm}[ii]) # mean rounding
#       W_q[ii] = xx[i] # floor rounding
#       W_q[ii] = xx[i +1] # ceil rounding
    return np.reshape(W_q, W.shape) * signs * M
```

**Fig.2.** Approximate code of discretization function. Commented lines show different variants of partition and different rounding methods

### 3. Discretization results for random numbers

We tested the discretization procedure on the two types of distributions, namely the Gaussian and the Laplacian distribution. In both cases we used 10 000 numbers from the given distribution to generate a vector and then performed the discretization using one of the abovementioned methods. We ranged the parameter $x_0$ from 0 to 1 and changed the number of bits from 2 to 6. The obtained results we collected in Tables 1 and 2 and showed them in Figures 3 and 4.

In Fig. 3 we show how the correlation value changes when we vary the parameter $x_0$. The maximal value of the correlation depends significantly on this parameter as well as on the discretization method. The values of maximums we collected in Table 1.

For comparison, we were able to implement an optimization algorithm providing an optimal partition into arbitrary intervals that did not use formulas (3) or (5). In Fig. 3, the horizontal point lines "optimal" correspond to the results obtained with the aid of this algorithm. Unfortunately, they define the optimal solution only if the number of the intervals is small (when $B$ equals to 2 or 3). When the number of the intervals is larger ($B \geq 4$) the algorithm often stucks and does not allow to obtain the optimal solution. In Fig. 3, we see that for $B = 5$ the *exponential_mean* method provides a slightly better solution then this algorithm.

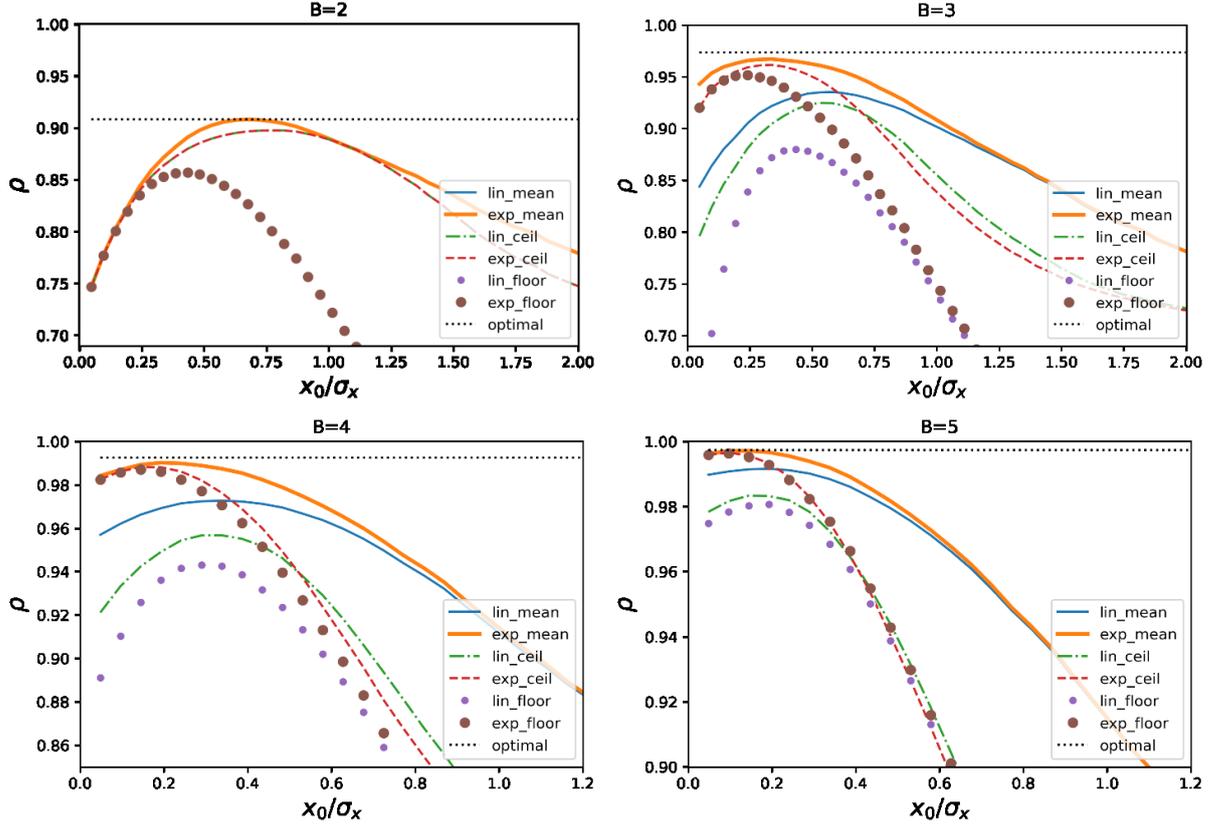

**Fig.3.** Dependence of correlation on $x_0$. On the horizontal axis is the value of $x_0$ normalized to the number standard deviation $\sigma_x$.

**Table 1.** Maximal correlation value when discretizing Laplacian random numbers.

| bits | Exponential mean | Exponential ceil | Exponential floor | Linear mean | Linear ceil | Linear floor |
|---|---|---|---|---|---|---|
| 2 | 0.9076±0.0006 | 0.8958±0.0041 | 0.8578±0.0008 | 0.9076±0.0006 | 0.8958±0.0041 | 0.8578±0.0008 |
| 3 | 0.9665±0.0017 | 0.9611±0.0018 | 0.9506±0.0025 | 0.9326±0.0051 | 0.9228±0.0033 | 0.8777±0.0062 |
| 4 | 0.99±0.0005 | 0.9882±0.0005 | 0.987±0.0006 | 0.9715±0.004 | 0.9557±0.0045 | 0.94±0.0075 |
| 5 | 0.9971±0.0001 | 0.9965±0.0001 | 0.9964±0.0002 | 0.9908±0.0016 | 0.9822±0.0027 | 0.9791±0.0035 |
| 6 | 0.99914±0.00004 | 0.99898±0.00004 | 0.99895±0.00004 | 0.9974±0.0005 | 0.9943±0.001 | 0.9938±0.0012 |

**Table 2.** Optimal values $x_0 / \sigma_x$ when discretizing Laplacian random numbers.

| bits | Exponential mean | Exponential ceil | Exponential floor | Linear mean | Linear ceil | Linear floor |
|---|---|---|---|---|---|---|
| 2 | 1.1272±0.0076 | 1.248±0.0382 | 0.7073±0.007 | 1.1272±0.0076 | 1.248±0.0382 | 0.7073±0.007 |
| 3 | 0.5778±0.0202 | 0.55±0.0168 | 0.4029±0.0106 | 0.9519±0.0421 | 0.9326±0.0489 | 0.7406±0.0142 |
| 4 | 0.3394±0.0118 | 0.2664±0.0077 | 0.2259±0.0066 | 0.5775±0.0441 | 0.5555±0.0425 | 0.5132±0.0341 |
| 5 | 0.2099±0.0091 | 0.1491±0.0051 | 0.1382±0.0055 | 0.3114±0.0274 | 0.2979±0.0267 | 0.2888±0.0251 |
| 6 | 0.1318±0.0078 | 0.0895±0.005 | 0.0854±0.0046 | 0.1603±0.0151 | 0.1544±0.0144 | 0.1528±0.0153 |

We found that the value $x_0$ fluctuates significantly from experiment to experiment. Next, we saw almost linear dependence between $x_0$ and the standard deviation of the numbers $\sigma_x$ (see Fig. 4). The fluctuations of the normalized value $x_0 / \sigma_x$ are much less (see Table 2). This value is exponentially dependent on the number of bits (see Fig. 5) and that is why we can obtain the following approximate formula for the optimal value of $x_0$:

$$x_0 \approx \frac{\sigma_x}{2^{B-2}} = \frac{2\sigma_x}{n} \qquad (8)$$

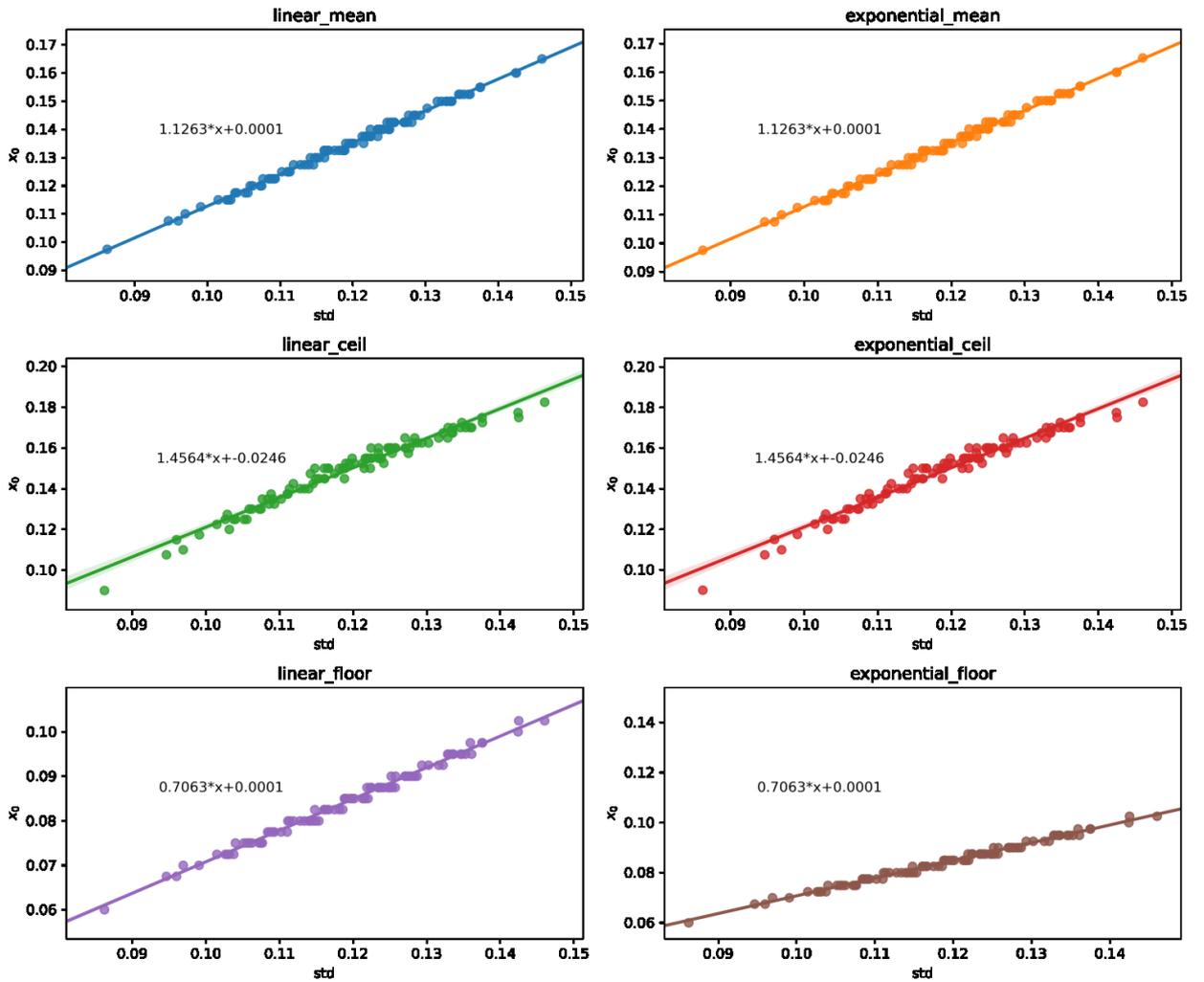

**Fig. 4.** Dependence of optimal $x_0$ on $\sigma_x$ when discretizing Laplace random numbers with $B = 2$. Each point corresponds to one experiment (100 experiments as a whole).

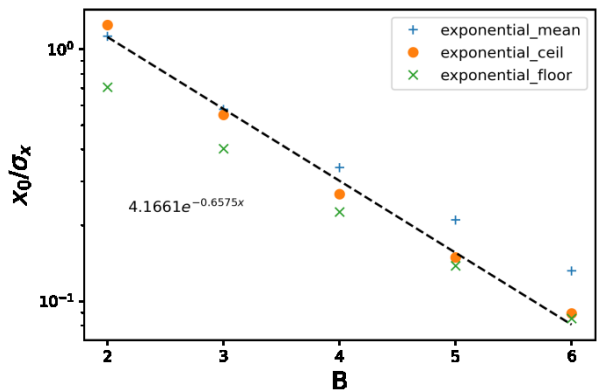

**Fig. 5.** Dependence of optimal value $x_0 / \sigma_x$ on number of bits $B$ for Laplace distribution. Dushed line is regression for exponential ceil method; regression equation is shone in Figure.

Our analysis allowed us to draw the following conclusions.

First, when the number of bits increases the correlation approaches 1; at that the approach speed depends exponentially on the number of bits (Table 1).

Second, the maximal correlation $\rho$ is achieved at a finite value of $x_0$. The right choice of $x_0$ allows us to reduce the difference $1-\rho$ significantly.

Third, the optimal value $x_0$ also depends on the number of bits significantly. It tends to zero exponentially when the number of bits increases (see Eq. (8)).

Finally, as we see from Table 1 the linear (uniform) discretization is 1 bit behind the exponential discretization. This means that the maximal correlations obtained in the framework of the linear partition is achieved by exponential partition when the number of bits is one bit less. Until now, we do not find the reason for this. However, this result means that the exponential discretization provides a one-bit gain.

### 3. Discretization of trained neural networks

We tested the above- described algorithms on neural networks pre-trained on the problem of recognition of photos from the database ImageNet [2]. The test set of photos of this database contains 50 000 pictures corresponding to 1000 classes. However, to reduce the computational complexity we used 1000 random images only and measured the classification accuracy prior and after discretization. Such tests allowed us to draw general conclusions about how the algorithm works.

For this analysis, we exploited ready for use models of neural networks from the library Keras Framework (tensorflow backend), namely VGG16, MobileNet-v2, Inception-v3, Xception, and ResNet-50 [11]. The work was done in the python language and the discretization procedure used the methods from the standard library numpy only.

When examining the trained neural networks such as VGG-16, ResNet-50, and so on, we found evident regularities in the distributions of the weights. For example, most frequently, the weight distributions are Gaussian or Laplacian and they are symmetrical with respect to zero. Probably this is a result of implementation of the regularizations L1 or L2 in the course of the training.

There are layer parameters that are not weights (for example, biases). We do not discretize them, because on the one hand, their number is not so large, and on the other hand, frequently their distributions differ from the Gaussian or Laplacian distributions.

Then the number of the weights we have to discretize decreases and we can estimate what percentage of the neural network will be discretized. For this purpose, we simply calculate the ratio of the weights we have to discretize to the total number of the weights. After excluding all the "unfitted" weights, we obtained the mobilenet-v2 neural network discretized by 95.7%. Other neural networks (Xception, Inceptionv3, ResNet-50) are discretized by 99%.

```
def quantize_whole_network(weights, discretization_func, layers_to_quantise, B):
    q_weights = weights.copy()
    for i in layers_to_quantise:
        $\sigma_W$ = weights[i].std()
        M = np.abs(weights[i]).max()
        x0 = $4\sigma_W$ / M /$2^B$
        qw = discretization_func(layer=weights[i], bitwidth=B, x0=x0)
        q_weights[i] = qw / qw.std() * $\sigma_W$
    return q_weights
```

**Fig. 6.** Approximate code for discretization of all neural network layers. As input argument *discretization_func* we use function shown in Fig. 2.

In Fig. 6, we show the procedure of the neural network discretization. Special attention should be paid to the next-to-last row. The point is that the maximization of the correlation is not the only one criterion of accurate discretization. We can multiply the discretized numbers by an arbitrary number without changing the correlation value. This was the reason why in our studies we additionally multiplied the discretized weights so that their dispersion remained the same prior and after the discretization.

As we see from Tables 3 and 4, when number of bits is equal to 6, the exponential discretization provides practically the same accuracy as the full precision weights. For some VGG-16 and ResNet50 neural networks, we obtain the comparable quality even using 4 bit exponential discretization. The same cannot be said for the linear discretization.

From Figs. 7 and 8 we see that the quality of the linear discretization is one and in some cases two bits behind the exponential discretization (the VGG-16 is an example).

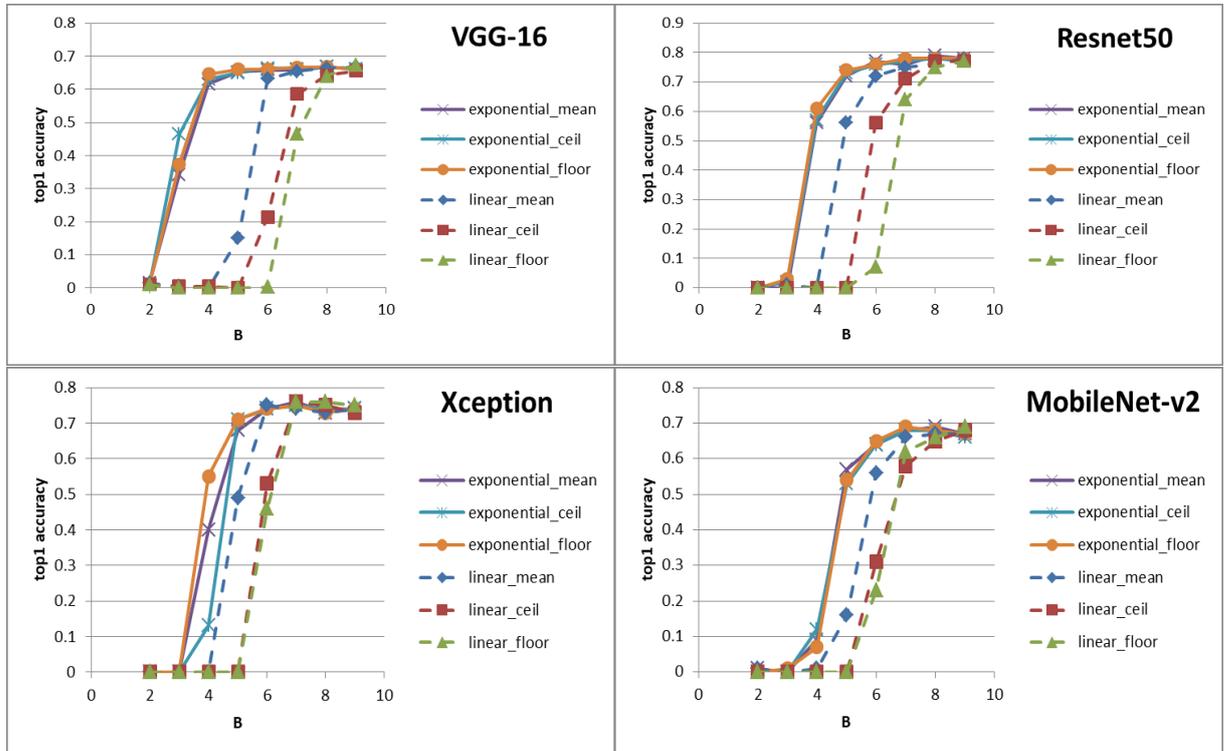

**Fig. 7.** Dependence of accuracy Top1 on number of bits of discretization for neural networks VGG-16, Resnet50, Xception, MobileNet-v2.

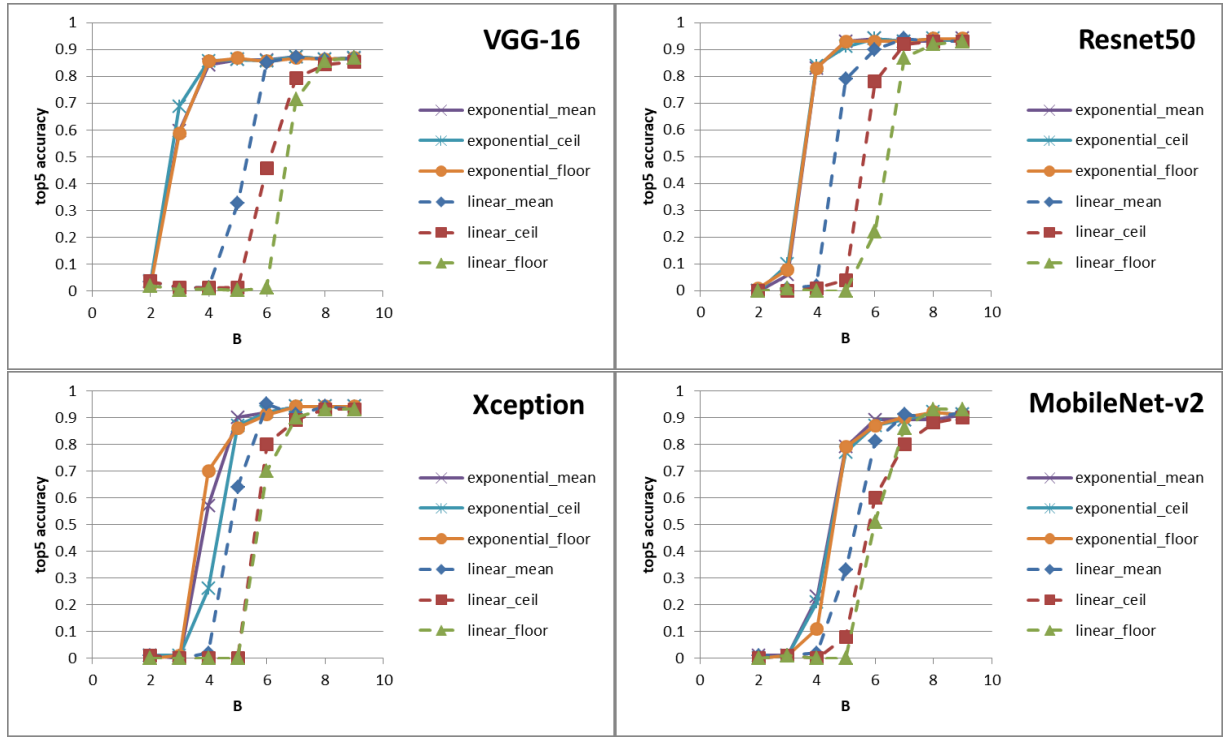

**Fig. 8.** Dependence of accuracy Top5 on number of bits of discretization for neural networks VGG-16, Resnet50, Xception, MobileNet-v2.

**Table 3.** Results of discretization of neural networks by means of *exponential_ceil* method

| Bits | top1 | | | | | top5 | | | | |
|---|---|---|---|---|---|---|---|---|---|---|
| | VGG-16 | Resnet50 | Xception | Inception-v3 | MobileNet-v2 | VGG-16 | Resnet50 | Xception | Inception-v3 | MobileNet-v2 |
| $n = 2$ (1+1 bit) | 0.01 | 0 | 0 | 0 | 0 | 0.04 | 0 | 0.01 | 0 | 0 |
| $n = 4$ (2+1 bit) | 0.46 | 0.02 | 0 | 0 | 0 | 0.69 | 0.1 | 0.01 | 0.01 | 0.01 |
| $n = 8$ (3+1 bit) | 0.63 | 0.57 | 0.13 | 0.26 | 0.12 | 0.87 | 0.84 | 0.26 | 0.35 | 0.21 |
| $n = 16$ (4+1 bit) | 0.68 | 0.73 | 0.71 | 0.70 | 0.53 | 0.89 | 0.91 | 0.87 | 0.9 | 0.77 |
| $n = 32$ (5+1 bit) | 0.69 | 0.76 | 0.74 | 0.75 | 0.67 | 0.9 | 0.94 | 0.92 | 0.91 | 0.87 |
| $n = 64$ (6+1 bit) | 0.7 | 0.77 | 0.75 | 0.76 | 0.69 | 0.9 | 0.93 | 0.94 | 0.91 | 0.89 |
| $n = 128$ (7+1 bit) | 0.7 | 0.78 | 0.74 | 0.77 | 0.7 | 0.9 | 0.94 | 0.94 | 0.91 | 0.89 |
| $n = 256$ (8+1 bit) | 0.7 | 0.77 | 0.74 | 0.77 | 0.71 | 0.9 | 0.93 | 0.94 | 0.93 | 0.9 |
| 32 bit | 0.703 | 0.757 | 0.796 | 0.777 | 0.708 | 0.909 | 0.933 | 0.945 | 0.93 | 0.896 |

**Table 4.** Results of discretization neural networks by means of *linear_mean* method

| Bits | top1 | | | | | top5 | | | | |
|---|---|---|---|---|---|---|---|---|---|---|
| | VGG-16 | Resnet50 | Xception | Inception-v3 | MobileNet-v2 | VGG-16 | Resnet50 | Xception | Inception-v3 | MobileNet-v2 |
| $n = 2$ (1+1 bit) | 0.00 | 0 | 0 | 0 | 0 | 0.00 | 0 | 0 | 0 | 0 |
| $n = 4$ (2+1 bit) | 0.00 | 0 | 0 | 0 | 0 | 0.01 | 0.01 | 0 | 0 | 0.01 |
| $n = 8$ (3+1 bit) | 0.00 | 0 | 0 | 0 | 0.01 | 0.01 | 0.02 | 0.02 | 0.03 | 0.02 |
| $n = 16$ (4+1 bit) | 0.15 | 0.56 | 0.49 | 0.07 | 0.16 | 0.33 | 0.79 | 0.64 | 0.13 | 0.33 |
| $n = 32$ (5+1 bit) | 0.69 | 0.72 | 0.75 | 0.7 | 0.57 | 0.89 | 0.9 | 0.95 | 0.88 | 0.81 |
| $n = 64$ (6+1 bit) | 0.69 | 0.75 | 0.74 | 0.75 | 0.67 | 0.9 | 0.94 | 0.91 | 0.93 | 0.91 |
| $n = 128$ (7+1 bit) | 0.7 | 0.76 | 0.73 | 0.76 | 0.7 | 0.9 | 0.93 | 0.94 | 0.93 | 0.9 |
| $n = 256$ (8+1 bit) | 0.7 | 0.78 | 0.74 | 0.77 | 0.71 | 0.91 | 0.93 | 0.94 | 0.93 | 0.9 |
| 32 bit | 0.703 | 0.757 | 0.796 | 0.777 | 0.708 | 0.909 | 0.933 | 0.945 | 0.93 | 0.896 |

## 4. Conclusions

We worked out the approach to the neural network discretization, which does not need the neural network retraining. The method does not demand sufficient computational resources but it provides a good recognition accuracy with 6 bit and sometimes, even with 4 bit (for example, in the case of VGG-16 neural network) weights. Consequently, it allows us to "lighten" the neural network approximately 5-8 times and in the same time to hold the classification accuracy.

**Acknowledgements**

The work was supported by Russian Foundation for Basic Research (RFBR Project 18-07-00750).